\documentclass[conference]{IEEEtran}
\usepackage{cite}
\usepackage{amsmath,amssymb,amsfonts}
\usepackage{algorithmic}
\usepackage{graphicx}
\usepackage{textcomp}
\usepackage{xcolor}
\usepackage{hyperref}
\usepackage{float}
\usepackage{microtype}
\usepackage[all]{nowidow}

\def\BibTeX{{\rm B\kern-.05em{\sc i\kern.025em b}\kern-.08em
    T\kern-.1667em\lower.7ex\hbox{E}\kern-.125emX}}

\makeatletter
\newcommand{\linebreakand}{%
  \end{@IEEEauthorhalign}
  \hfill\mbox{}\par
  \mbox{}\hfill\begin{@IEEEauthorhalign}
}
\makeatother

\begin{document}

\title{Copyright Detection in Large Language Models: 

An Ethical Approach to Generative AI Development}

\author{

\IEEEauthorblockN{David Szczecina}
\IEEEauthorblockA{
    \textit{University of Waterloo} \\
    david.szczecina@uwaterloo.ca
}

\and
\IEEEauthorblockN{Senan Gaffori}
\IEEEauthorblockA{
    \textit{University of Waterloo} \\
    senan.gaffori@uwaterloo.ca
}

\and
\IEEEauthorblockN{Edmond Li}
\IEEEauthorblockA{
    \textit{University of Waterloo} \\
    e26li@uwaterloo.com
}

} 
\maketitle

\begin{abstract}
The widespread use of Large Language Models (LLMs) raises critical concerns regarding the unauthorized inclusion of copyrighted content in training data. Existing detection frameworks, such as DE-COP, are computationally intensive, and largely inaccessible to independent creators. As legal scrutiny increases, there is a pressing need for a scalable, transparent, and user-friendly solution. This paper introduce an open-source copyright detection platform that enables content creators to verify whether their work was used in LLM training datasets. Our approach enhances existing methodologies by facilitating ease of use, improving similarity detection, optimizing dataset validation, and reducing computational overhead by 10-30\% with efficient API calls. With an intuitive user interface and scalable backend, this framework contributes to increasing transparency in AI development and ethical compliance, facilitating the foundation for further research in  responsible AI development and copyright enforcement.
\end{abstract}

\section{Introduction}

\subsection{Motivation}

Large Language Models (LLMs) such as GPT-4 and Claude have revolutionized natural language processing, but also raise legal and ethical concerns about the unauthorized use of copyrighted content in training datasets\cite{respect_copyright}. Proprietary models often rely on large-scale web scraping\cite{decop}, incorporating copyrighted material without clear consent mechanisms, compensation, and intellectual property protection\cite{rag_copyright}.

A major concern is the lack of compensation for content creators whose work is used without permission. Legal frameworks for AI copyright enforcement are rapidly evolving, with landmark cases like New York Times v. OpenAI \cite{debate} bringing increased scrutiny to dataset curation. Transparency in AI training datasets is essential to ensure responsible and ethical development. Research indicates that as models increase in size, memorization tendencies become more pronounced, particularly in models exceeding 100 billion parameters \cite{debate}, increasing the risk of unauthorized reproduction of copyrighted content.

Current detection methods, such as plagiarism checkers and statistical techniques, struggle to identify subtly paraphrased copyrighted content\cite{decop} \cite{info_probing}. While frameworks such as DE-COP offer promising approaches, they remain computationally expensive and complex; making them impractical for independent creators and smaller organizations. A scalable, cost-effective, and user-friendly solution is needed to verify whether copyrighted works have been used in LLM training datasets.

\begin{figure}[H] 
\vspace{-10pt}
\centering
\includegraphics[width=\columnwidth]{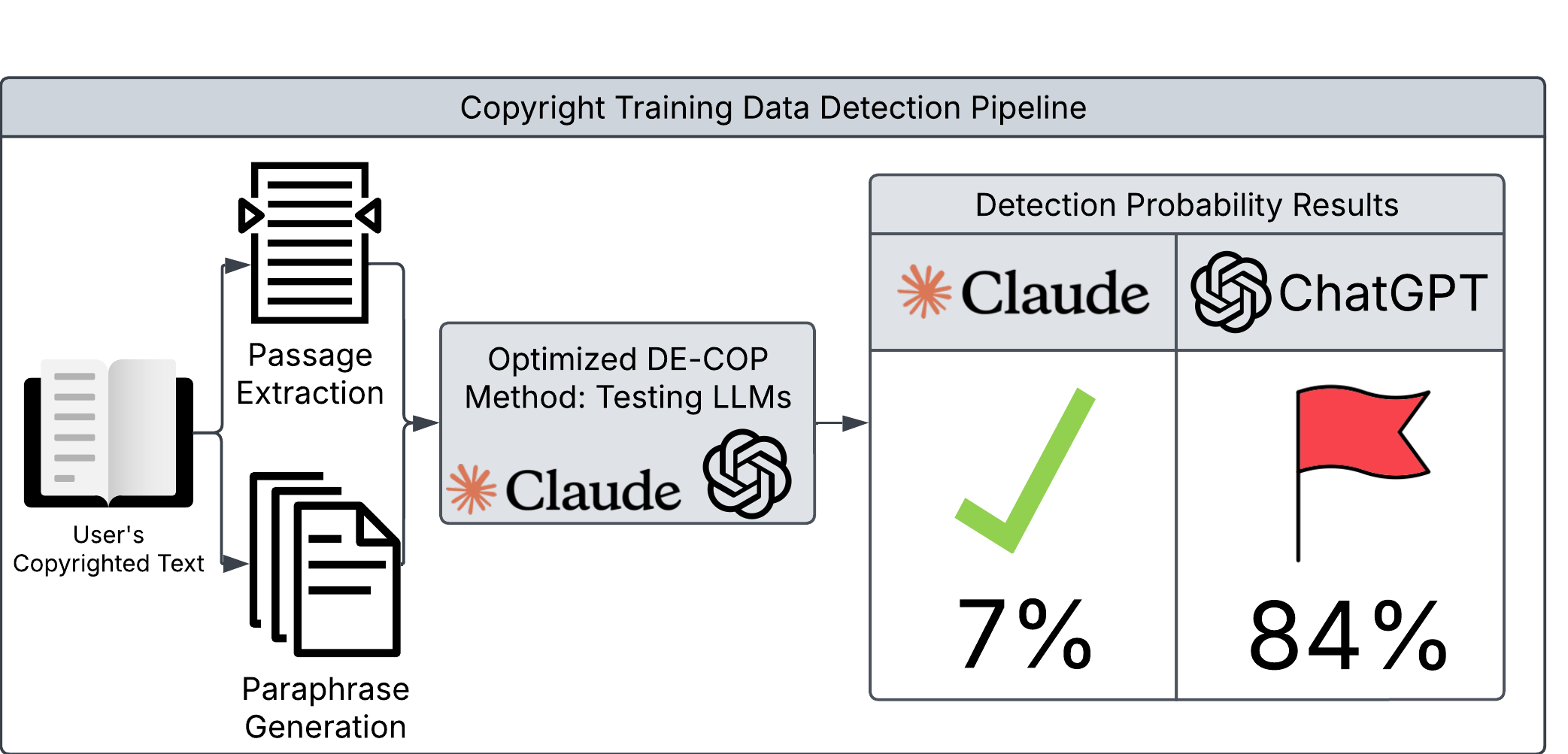}
\vspace{-15pt} 
\caption{Unique passages are extracted and paraphrased from a users content, next an LLM is prompted to determine the original passage. Final scores show the probability of the copyrighted content being used in training the LLM}
\label{fig:Hiigh Level Overview}
\end{figure}

\subsection{Related Works}

The detection of copyrighted content in LLM training datasets has been the subject of increasing research attention, particularly as legal and ethical concerns surrounding dataset curation intensify. While traditional plagiarism detection tools struggle to identify AI-generated reproductions of proprietary content \cite{decop}, several machine learning-based approaches have been proposed to address this issue.

Membership inference attacks \cite{plagiarism} analyze a model’s confidence scores to determine whether a given text sample was likely included in the training data. Although effective in controlled experiments, this approach requires adversarial access to the model and often produces inconclusive results due to dataset augmentation and model fine-tuning techniques.
Similarly, perplexity-based analysis is another detection approach by evaluating how confidently an LLM predicts a passage of text \cite{confidence}. Low perplexity scores suggest memorization, however, this method struggles to distinguish between legally sourced and unauthorized content, making it unreliable for copyright enforcement.
Another proposed approach is digital watermarking \cite{watermark}, where imperceptible markers are embedded into text data before model training. While useful for tracking known copyrighted works, watermarking is ineffective against existing datasets that were scraped from the web and fails to detect content that has been paraphrased or restructured.

\begin{figure}[H] 

\centering
\includegraphics[width=\columnwidth]{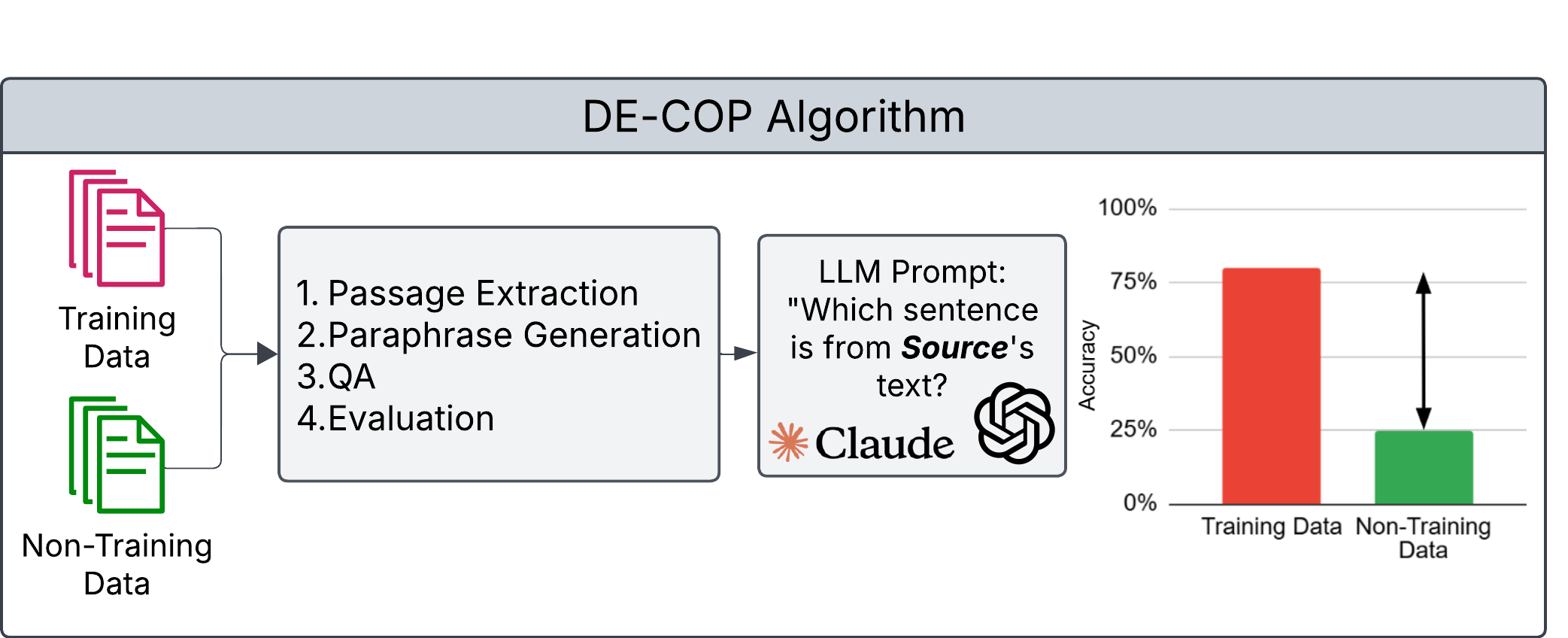}
\vspace{-15pt} 
\caption{DE-COP System Overview}
\label{fig:DE-COP System Overview}
\end{figure}

A more recent approach, DE-COP: Detecting Copyrighted Content in Language Models Training Data, \cite{decop}, introduces a method to determine whether a language model has memorized copyrighted content. Unlike statistical approaches, DE-COP introduces a multiple-choice question-answering framework, where an LLM must distinguish an original verbatim passage from paraphrased alternatives. If a model consistently selects the correct passage, this suggests that the content was likely included in its training data. An overview of the DE-COP system is illustrated in Figure~\ref{fig:DE-COP System Overview}.
Despite its advantages, DE-COP is computationally expensive, requiring approximately 590 seconds per book for open-source models (LLaMA-2 70B) \cite{Llama} and 331 seconds on ChatGPT \cite{decop} \cite{gpt}. Methods such as Min-K\%-Prob\cite{confidence}, Prefix Probing\cite{prefix_probing} and Name Cloze Task \cite{name_cloze} only required 13-17 seconds to perform the same task \cite{decop}. Additionally, the datasets presented in DE-COP were found to contain NULL values, errors message outputs, half finished sentences, and new paraphrases ranged from being 20\% to 250\% as long as the original passage \cite{decop}. DE-COP lacks robust features to handle these errors in its own dataset, and its evaluation metrics were based on questionable data, leaving lots of room for improvements.

\vspace{-10pt}

While previous methods provide partial solutions to the problem of detecting copyrighted content in LLM training data, they often fall short in generalization and effectiveness. DE-COP introduced a black-box-compatible alternative that significantly improves detection accuracy \cite{decop}. However, optimizing its computational efficiency and reducing selection biases remains an open challenge for future work.

\subsection{Problem Definition}

Despite the concerns for copyright material in LLM training data, existing copyright detection methods remain insufficient and inaccessible. Traditional plagiarism detection tools struggle to identify paraphrased or subtly modified copyrighted content, making enforcement difficult \cite{decop}. Additionally, computationally intensive frameworks such as DE-COP, are impractical for independent content creators due to their technical complexity and high computational costs \cite{energy_costs}. The absence of cost-effective and user-friendly solutions further limits the ability to verify whether copyrighted works have been used in LLM training.
This paper introduces an open-source framework that enhances dataset validation, improving similarity detection, and optimizes computational efficiency. By significantly reducing processing costs while maintaining detection accuracy, our approach provides a scalable and accessible platform for copyright verification. This initiative  ensures AI transparency, promotes fair compensation for content creators, and supports ethical AI development in the rapidly evolving landscape of generative~AI.

\section{Methodology}

\begin{figure}[H] 
\centering
\includegraphics[width=\columnwidth]{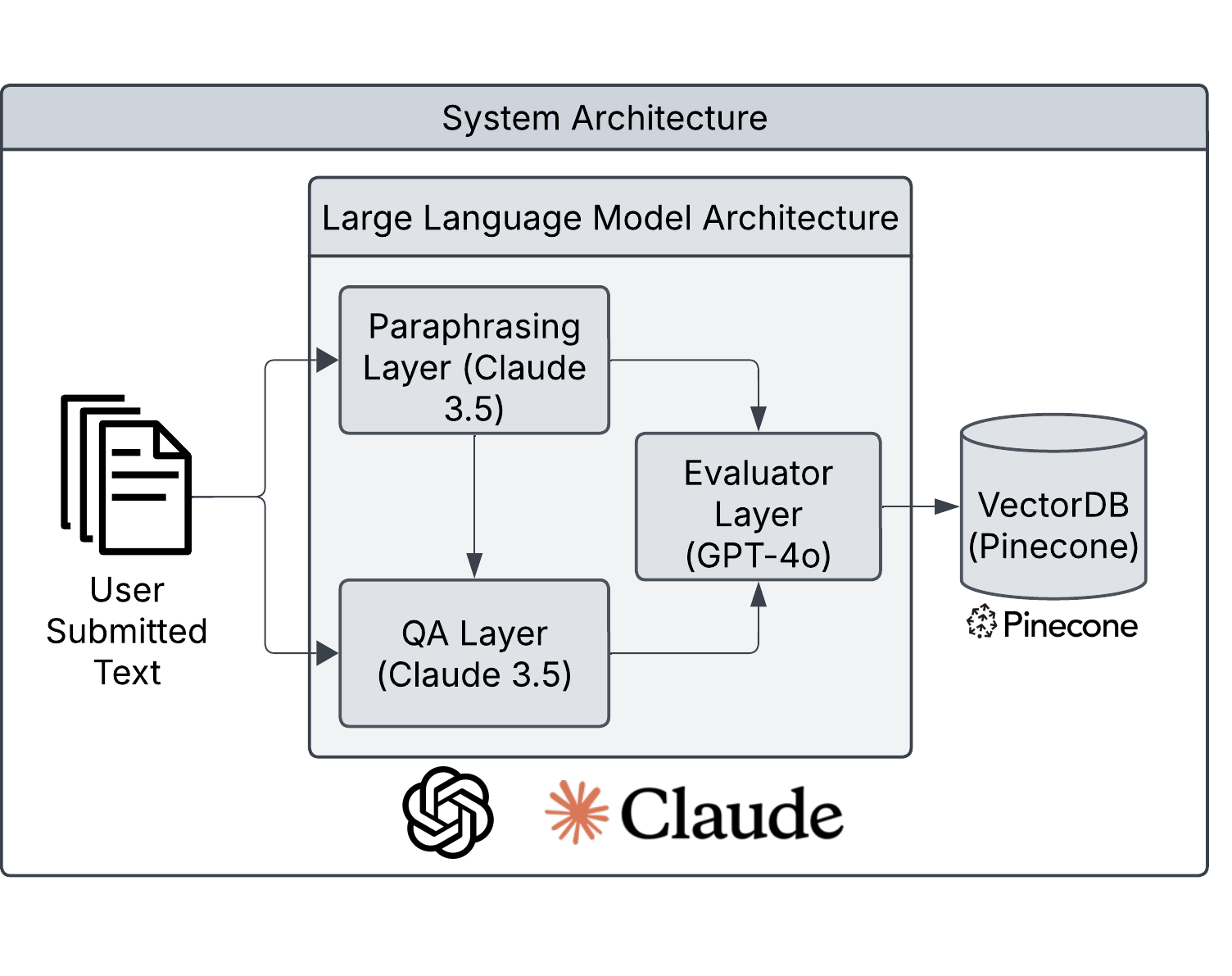}
\vspace{-35pt} 
\caption{System Architecture Diagram}
\label{fig:system_archetecture}
\end{figure}

This project features a web-based UI where users can submit content for evaluation. The backend evaluation system, runs a multi-layered evaluation workflow integrating passage extraction, paraphrase generation, question-answering, multiple-choice evaluation, and statistical analysis to detect copyrighted content in LLM training data. A vector store maintains a record of previously evaluated content, allowing the system to check for duplicates, avoiding redundant evaluations, as well as search through past evaluations. Users can access a dashboard and analytics page to view evaluation histories and check accuracy metrics. An overview of the system architecture is illustrated in Figure~\ref{fig:system_archetecture}.

\vspace{-10pt}

\subsection{Passage Extraction}
Selecting highly unique passages enhances the accuracy and effectiveness of detecting copyrighted content in language model training data\cite{wynter}. Unique passages minimize the risk of incorporating common phrases and generic text that may not sufficiently challenge the model's memorization capabilities\cite{memoization}. To identify these passages, the BM25 algorithm \cite{bm25} was employed to vectorize passages within the document and calculate similarity scores between them. By treating each passage as a query against the entire document, BM25 assigned scores based on term frequency and inverse document frequency. Passages with the lowest BM25 scores, indicating minimal similarity to other passages, were considered the most unique. These high-uniqueness passages were prioritized for use in the evaluation layer, providing a robust foundation for detecting memorized content.

\subsection{Paraphrase Generation}

The paraphrase generation layer is implemented using LangGraph's StateGraph to create a modular and dynamic workflow. This layer utilizes the Claude 3.5 Sonnet model~\cite{claude} via the ChatAnthropic API with a temperature setting of 0.7, ensuring a balance between creativity and control in generating paraphrases. Unlike the original DE-COP approach, which applies standard paraphrasing prompts, our method introduces specific paraphrasing strategies, including passive voice conversions, question-based restructuring, and language simplification. These templates promote greater diversity in paraphrases \cite{bandelquality}, enhancing the robustness of the evaluation by reducing model prediction patterns. Additionally, the implementation proposes XML formatting for paraphrases to support integration with instructional models, offering improved compatibility and structured data handling not present in the original DE-COP~method.

\subsection{Question-Answering}

The QA layer is also built using LangGraph’s StateGraph, facilitating an automated workflow that handles both ``create'' and ``format'' modes for generating evaluation questions. While the original DE-COP\cite{decop} method primarily focused on generating standardized multiple-choice questions, our implementation expands functionality by allowing the creation of custom questions that use exact text from the input content. The QA layer uses the ChatAnthropic model to generate questions in a structured JSON format, improving downstream processing and maintaining output consistency. This added flexibility enhances the evaluation's accuracy by testing the model's memorization across varied question formats, contributing to a more thorough assessment of the model’s exposure to copyrighted content.

\subsection{Multiple-Choice}
The multiple-choice layer employs LangGraph to manage answer selection and evaluation workflows. In contrast to the original DE-COP's\cite{decop} exhaustive approach of generating all permutations of answer choices, our initial implementation used a simplified randomization strategy to prevent selection bias. However, we propose an enhancement that includes a dedicated permutation function to fully automate all possible answer orderings within LangGraph. The evaluation prompts are designed to elicit concise, formatted responses from the model, minimizing noise and ensuring clarity in the output. Incorporating full permutation handling would better strengthen the mitigation of selection biases in model responses.

\subsection{Evaluation}
The evaluation layer integrates multiple components, including paraphrase generation, question answering, multiple-choice testing, and statistical analysis, using GPT-4o \cite{gpt} via LangGraph. Our implementation extends upon DE-COP's framework by incorporating advanced statistical methodologies such as receiver-operating characteristic (ROC) curve analysis, area under the curve (AUC) scoring, and hypothesis testing. This layer provides deeper insights into performance through robust statistical methods. A key enhancement over previous methodologies is the introduction of a permutation function which generates all answer permutations; mitigating selection biases in LLMs. The evaluation prompts guide the model through a structured evaluation process, emphasizing precise and formatted responses. These enhancements create a more modular and statistically robust framework, improving the accuracy and reliability of detecting copyrighted content in LLM training data.

\subsection{Logging System and Similarity Search}

To enable content tracking and retrieval, the system incorporates Pinecone, a serverless vector database. Documents are embedded using all-MiniLM-L6-v2 \cite{minilm} (embedding model from HuggingFace) which offers a strong balance of embedding quality and efficiency. The model generates 384-dimensional embeddings to support fast and accurate approximate nearest neighbour (ANN) searches, while integrating seamlessly with Pinecone and LangGraph. Metadata attributes such as copyright ownership, evaluation timestamps, evaluation results, and content type, are stored directly in Pinecone as key-value pairs. Logging metedata enables quick access and tractability during content evaluations without requiring an external database. 
The ingestion pipeline is designed for single-document processing, embedding each submission and storing it with a unique identifier. To evaluate content, the system compares new submissions against stored vectors, retrieving the most similar documents and their metadata. This streamlined vectorized approach supports the goal of creating an open-source API that logs copyrighted content appearing in LLM training data; promoting transparency and accountability in AI development.

\subsection{Data Processing Improvements}

An analysis of DE-COP’s dataset revealed several inconsistencies such as NULL values, API output errors, inconsistent formatting, and extreme variations in passage length\cite{decop}. These inconsistencies negatively impacted accuracy, skewed model predictions and increased token usage by up to 50\%. To address this, a preprocessing pipeline was implemented using SBERT embeddings \cite{sbert} and cosine similarity, ensuring that paraphrases retain semantic integrity, and any invalid passages are filtered out. Additionally, passage lengths are normalized to prevent instances where paraphrases are excessively short or long, improving paraphrase consistency. These enhancements eliminated inconsistencies in DE-COP’s dataset\cite{decop}, providing results that are more reproducible and statistically sound.

To further reduce API costs, the multiple-choice selection was expanded from three to four paraphrased options, reducing the probability of the original passage being randomly selected by 20\%. By decreasing the probability of Type I error, the total number of passages requiring evaluation can be decreased without compromising the experiment's statistical power or significance. Consequently, this optimization substantially lowers overall API consumption by requiring less passages to be evaluated.

\section{Results}
Our proposed framework demonstrates significant improvements in detection accuracy, computational efficiency, and accessibility over existing methodologies. By providing our open-source solution as a hosted platform, we remove technical barriers, promoting ease of use and access to individual content creators.

The multi-layered workflow, which integrates passage extraction, paraphrase generation, question-answering, and multiple-choice testing, effectively differentiates between memorized (copyrighted) and non-memorized text. By integrating a pre-screening pipeline using SBERT embeddings, cosine similarity, and normalized passage lengths, errors are caught and filtered out; enhancing reproducibility. The multiple-choice evaluation layer, with a streamlined randomization strategy and restructuring of question format reduced API consumption by 10-30\%. Additionally, the Pinecone vector store  enhances scalability and duplicate detection, avoiding redundant evaluations. These enhancements provide a scalable and practical solution that outperforms existing approaches, such as DE-COP, supporting ethical AI development and fair compensation for content creators.

\section{Conclusion}

This paper introduces an open-source framework for detecting copyrighted content in LLM training datasets, addressing key limitations in accessibility, detection accuracy, and cost efficiency found in previous approaches such as DE-COP. By enhancing similarity detection, refining dataset validation, and optimizing computational efficiency, our system provides a scalable and accessible solution for copyright verification.
Through our user-friendly interface, content creators can easily determine whether their work was appropriated for AI development, without a high technical barrier to entry. 
By promoting transparency and encouraging accountability, our system ultimately paves the way for ethical AI development.

\section{Future Work}

Future research may focus on developing methods for selective knowledge removal, such as \textit{Unlearn} \cite{unlearn}, to enable the erasure of copyrighted content from LLMs. This knowledge removal technique could be implemented for some of the standard LLM pretraining datasets such as \textit{C4} \cite{c4} and \textit{Pile} \cite{pile}. The legal implications of dataset memorization also warrant further investigation, particularly as AI copyright regulations continue to evolve. Additionally, expanding the scalability and adoption of our platform across different AI models and regulatory frameworks will be crucial for broader impact.

\section*{Acknowledgments}

This research was enabled in part by funding, resources, and support provided by Wat.AI, the Waterloo AI Institute, and the Sedra Student Design Centre.

\bibliographystyle{IEEEtran}
\bibliography{references}

\end{document}